\documentclass{article}

\usepackage[preprint]{neurips_2026}

\usepackage[utf8]{inputenc}
\usepackage[T1]{fontenc}
\usepackage{hyperref}
\usepackage{url}
\usepackage{booktabs}
\usepackage{amsmath}
\usepackage{amsfonts}
\usepackage{amssymb}
\usepackage{graphicx}
\usepackage{tabularx}
\usepackage{float}
\usepackage{tikz}
\usetikzlibrary{positioning, arrows.meta, shapes.geometric, fit, backgrounds}
\usepackage{pgfplots}
\usepackage{subcaption}
\usepackage{xcolor}
\setcitestyle{numbers}
\pgfplotsset{compat=1.18}

\title{Evaluating Temporal Semantic Caching and Workflow Optimization in Agentic Plan-Execute Pipelines}

\author{
  Alimurtaza Merchant\thanks{Equal contribution.} \\
  Columbia University \\
  \texttt{amm2640@columbia.edu} \\
  \And
  Krish Veera\footnotemark[1] \\
  Columbia University \\
  \texttt{krv2123@columbia.edu} \\
  \And
  Sajal Kumar Goyla\footnotemark[1] \\
  Columbia University \\
  \texttt{sg4607@columbia.edu} \\
  \And
  Shambhawi Bhure\footnotemark[1] \\
  Columbia University \\
  \texttt{sb5185@columbia.edu} \\
  \And
  Dhaval Patel \\
  IBM \\
  \texttt{pateldha@us.ibm.com} \\
  \And
  Kaoutar El Maghraoui \\
  IBM Research \\
  \texttt{kelmaghr@us.ibm.com} \\
}

\begin{document}

\maketitle

\begin{abstract}

Industrial asset operations workflows are latency-sensitive because a single user query may require coordination over sensor data, work orders, failure modes, forecasting tools, and domain-specific agents. We evaluate this problem on AssetOpsBench (AOB), an industrial agent benchmark whose plan-execute pipeline exposes repeated overhead from tool discovery, LLM planning, MCP tool execution, and final summarization. Existing LLM caching techniques such as KV-cache reuse and embedding-based semantic caching were designed for chatbot serving and break down when output validity depends on time, asset, or sensor parameters. We propose two complementary optimization layers for AOB plan-execute pipelines: a temporal semantic cache and a set of MCP workflow optimizations combining disk-backed tool-discovery caching and dependency-aware parallel step execution. MCP workflow optimizations corresponded to a 1.67× speedup and reduced median end-to-end latency by about 40.0\% while the temporal-cache benchmark achieved a median of 30.6× speedup on cache hits. Beyond the speedup, our results expose a concrete failure mode of pure semantic caching for parameter-rich industrial queries, providing a critical analysis of how caching choices interact with evaluation correctness in MCP-backed agent benchmarks.
\end{abstract}

\section{Introduction}
\label{sec:intro}

LLM-based agents increasingly serve as orchestration layers over domain-specific tools and data sources~\citep{yao2023react, schick2023toolformer, patil2024gorilla}. Many such agents follow the Plan-Execute paradigm: a planner LLM decomposes a user query into a sequence of tool calls, and an executor invokes those tools, often through standardized interfaces such as the Model Context Protocol (MCP)~\citep{anthropic2024mcp}. While expressive, this two-stage structure introduces substantial wall-clock latency. A single query can require tool discovery, multi-step planning, multiple MCP tool invocations, and a final summarization pass before an answer is returned.

This latency is particularly acute in industrial asset operations, where queries naturally span heterogeneous data sources: sensor telemetry, work orders, failure modes, and time-series forecasts. The AssetOpsBench benchmark~\citep{aob2024} formalizes this setting with four MCP-backed domain servers and a corpus of human-authored operational queries. In practice, the same operator may issue many semantically related queries against the same assets: paraphrases, repetitions, parameter shifts (Chiller~6 vs. Chiller~9), or time-window shifts (yesterday vs. last week). A naive plan-execute implementation pays the full orchestration cost on every query, and at paraphrase scale this makes systematic evaluation of MCP-backed agents prohibitively slow.

Caching is the standard remedy for repeated computation in LLM serving, but existing techniques were designed for chatbot workloads. Context (KV) caching~\citep{gim2024promptcache, yao2025cacheblend, jin2024ragcache, liu2024cachegen} reuses prefill states for identical prefixes; semantic caching~\citep{bang2023gptcache, schroeder2025vectorq} reuses (input, output) pairs across paraphrases via embedding similarity. Neither approach matches the structure of industrial agent queries, where output validity depends on external state (asset, sensor, time window) that is not visible in the query embedding. Recent work on Agentic Plan Caching~\citep{zhang2025apc} addresses part of this gap by caching plan templates rather than answers, but does not address temporal validity, which is central in industrial telemetry settings where ``what happened yesterday'' resolves to a different window each day.

\paragraph{Our approach.}
We propose two optimization layers for AOB plan-execute pipelines. At the query level, we build a \textit{temporal semantic cache} with a lightweight temporal classifier that routes each query into one of four buckets: Volatile (live state, bypass cache), Static (no temporal dependence, standard semantic match), Relative (e.g., ``yesterday,'' resolved into a concrete window), or Anchored (fixed time window, matched against compatible windows). Static and (resolved) Anchored queries enter embedding-based retrieval followed by a reranker-based judge. At the workflow level, we add two \textit{MCP optimizations}: disk-backed tool-discovery caching and dependency-aware parallel step execution over a persistent server pool. The two layers are independently beneficial and additive: the MCP layer reduces latency on every query regardless of cache state, while the cache layer adds large further savings when a query resolves to a valid hit.

\paragraph{Contributions.}
\begin{enumerate}
\item \textbf{Temporal semantic caching for industrial agents.} We extend semantic caching with a pre-retrieval temporal classifier and a window-aware judger that addresses the parameter-and-time sensitivity of industrial queries.
\item \textbf{MCP workflow optimizations.} We combine discovery-phase caching, DAG-layered parallel execution, and a persistent server pool to reduce per-query orchestration overhead in MCP-backed plan-execute pipelines.
\item \textbf{Refined evaluation setup for AOB.} We provide a paired baseline-vs-optimized harness, a paraphrase-tier test set with parent-id-based ground truth for cache hit/miss labelling, and per-phase latency profiling that makes systematic AOB ablations tractable on a single machine.
\item \textbf{Critical analysis of caching as an evaluation choice.} We expose a concrete failure mode: pure semantic similarity is not a sound proxy for answer validity in parameter-rich AOB queries, capping hit-decision F1 near 0.67 in our setting. This gives a measurable handle on when caching is safe to use as part of an evaluation pipeline, and when it is not.
\end{enumerate}

\section{Background and Motivation}
\label{sec:background}

\subsection{AssetOpsBench and MCP-backed Plan-Execute Agents}

AssetOpsBench (AOB)~\citep{aob2024} is a benchmark for evaluating LLM agents on industrial asset operations and maintenance workflows. AOB exposes four specialized domain servers covering Internet of Things (IoT) telemetry, Failure Mode and Sensor Relation (FMSR), Time Series Foundation Models (TSFM)~\citep{das2024timesfm}, and Work Order (WO) records, all wrapped under the Model Context Protocol (MCP)~\citep{anthropic2024mcp}. The benchmark scenarios are written as human-authored natural-language operational queries rather than database-level API calls, reflecting the language an operator or reliability engineer would actually use.

\begin{figure}[ht]
    \centering
    \includegraphics[width=0.78\linewidth, height=0.20\textheight, keepaspectratio]{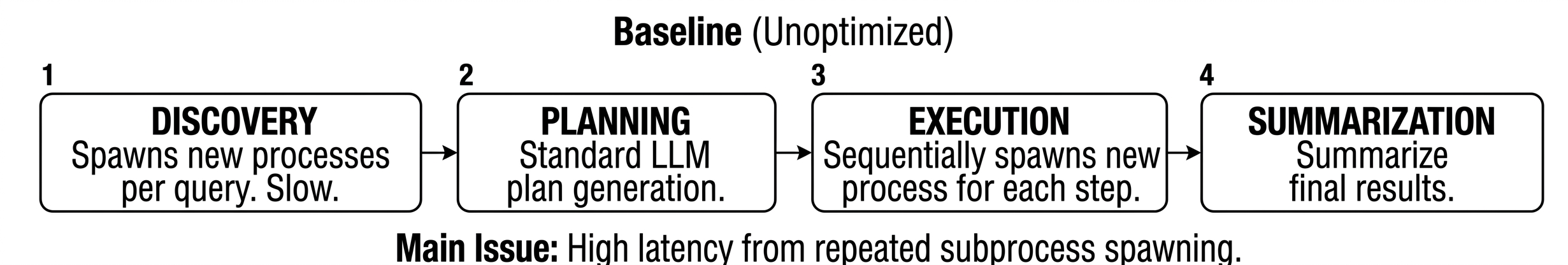}
    \caption{MCP Workflow.}
    \label{fig:mcp-workflow}
\end{figure}

In a Plan-Execute pipeline~\citep{yao2023react}, a query is not answered by a single LLM call. Instead, the workflow decomposes into four phases: \textit{Discovery}, which spawns MCP servers and collects tool signatures via \texttt{list\_tools()}; \textit{Planning}, which uses an LLM to convert the query and tool catalog into a structured plan; \textit{Execution}, which resolves tool arguments and invokes the relevant MCP tools; and \textit{Summarization}, which uses an LLM to synthesize the tool outputs into a response. Figure~\ref{fig:mcp-workflow} illustrates this structure and illustrates the baseline path.

The Plan-Execute abstraction is useful because it exposes a structured plan before tool execution begins. However, this separation does not automatically imply parallelism: many implementations consume the generated plan strictly sequentially. The optimization opportunity comes from treating the plan as a directed acyclic graph and dispatching dependency-independent steps concurrently, while preserving order across true dependencies.

\subsection{LLM Caching for Agents: Methods and Limitations}\label{sec:caching-limits}

Caching is one of the most widely adopted techniques for reducing LLM serving cost. \textit{Context caching}~\citep{gim2024promptcache, yao2025cacheblend, jin2024ragcache, liu2024cachegen} stores key-value pairs from the prefill phase and reuses them when prompt prefixes recur. \textit{Semantic caching}~\citep{bang2023gptcache, schroeder2025vectorq} stores (input, output) pairs and matches new queries by embedding similarity, exploiting the fact that paraphrases share underlying intent. \textit{Plan caching}~\citep{zhang2025apc} caches plan templates extracted from completed agent executions and adapts them to new queries with a lightweight model. We find that all three families have limitations specific to MCP-backed industrial agent benchmarks like AOB:

\textbf{1) Static-Output Assumption.} Semantic caching assumes that outputs depend only on the input prompt~\citep{bang2023gptcache, schroeder2025vectorq}. This holds for chatbots but fails in AOB, where outputs depend on external state queried at run time. For example, ``what is the status of work order WO-1234'' returns a different answer depending on whether the order is open or closed, but the query text and its embedding are identical each time. A cache that keys on the input alone cannot detect that the stored answer is stale.

\textbf{2) Parameter Insensitivity.} Embedding-based retrieval captures the linguistic structure of a query but is insensitive to its operational parameters. ``Failure modes detectable by Chiller~6 Efficiency sensor'' and ``Failure modes detectable by Chiller~9 Efficiency sensor'' embed close together because they share the same sentence frame, yet they require disjoint tool calls and produce disjoint answers. Threshold tuning trades false positives against false negatives but does not eliminate this structural mismatch between what the embedding encodes and what determines a correct answer~\citep{schroeder2025vectorq}.

\textbf{3) Temporal Blindness.} Many AOB queries contain explicit or relative time expressions: ``last week,'' ``yesterday,'' ``the past 24 hours.'' Pure embedding similarity treats two such queries as equivalent regardless of their resolved windows, which is incorrect when the underlying telemetry has changed. Existing semantic caching frameworks~\citep{bang2023gptcache, schroeder2025vectorq} do not expose a mechanism to distinguish queries that differ only in their resolved temporal anchor.

These limitations motivate a temporal-aware cache that distinguishes \textit{semantic relatedness} from \textit{safe answer reuse}, combined with workflow-level optimizations that reduce per-query overhead even on cache misses.

\subsection{Related Work}

\paragraph{Agent memory and plan reuse.}
Several recent systems augment LLM agents with external memory~\citep{packer2023memgpt, xu2025amem, sumers2023cognitive}. Agent Workflow Memory~\citep{wang2024awm} extracts and reuses workflow patterns to improve task success rates. Asteria~\citep{asteria2025} provides the semantic-cache primitives we build on (ANN over query embeddings, a reranker-based judger, LCFU eviction, and Markov prefetching) for general agentic LLM tool access. Agentic Plan Caching~\citep{zhang2025apc} extends agent-side reuse to a serving-cost objective by caching plan templates and adapting them with a small LM. Our work differs in two ways: we target MCP-backed industrial benchmarks where temporal validity is central, and we add a temporal classification layer in front of Asteria-style retrieval to handle relative time expressions and live-state queries.

\paragraph{LLM serving infrastructure.}
Modern LLM serving systems such as vLLM~\citep{kwon2023vllm} and SGLang~\citep{zheng2024sglang} optimize inference at the engine level through KV-cache management and structured generation. Our optimizations sit one layer up: they target the agent orchestration loop and are compatible with any underlying serving engine.

\paragraph{Multi-agent orchestration and benchmarks.}
Multi-agent collaboration has been studied in systems such as Mixture-of-Agents~\citep{wang2024moa} and surveyed broadly in~\citep{guo2024mas, park2023generativeagents}. Agent benchmarks like GAIA~\citep{mialon2024gaia} and Minions~\citep{narayan2025minions} evaluate task success and cost; AOB~\citep{aob2024} extends this to industrial operations with MCP-exposed tooling. Our contribution is a refinement of the AOB evaluation setup itself, making at-scale paraphrase-tier ablations practical.

\section{The Optimization Framework}
\label{sec:framework}

\subsection{Overview}

\begin{figure}[ht]
    \centering
    \includegraphics[width=0.78\linewidth, height=0.20\textheight, keepaspectratio]{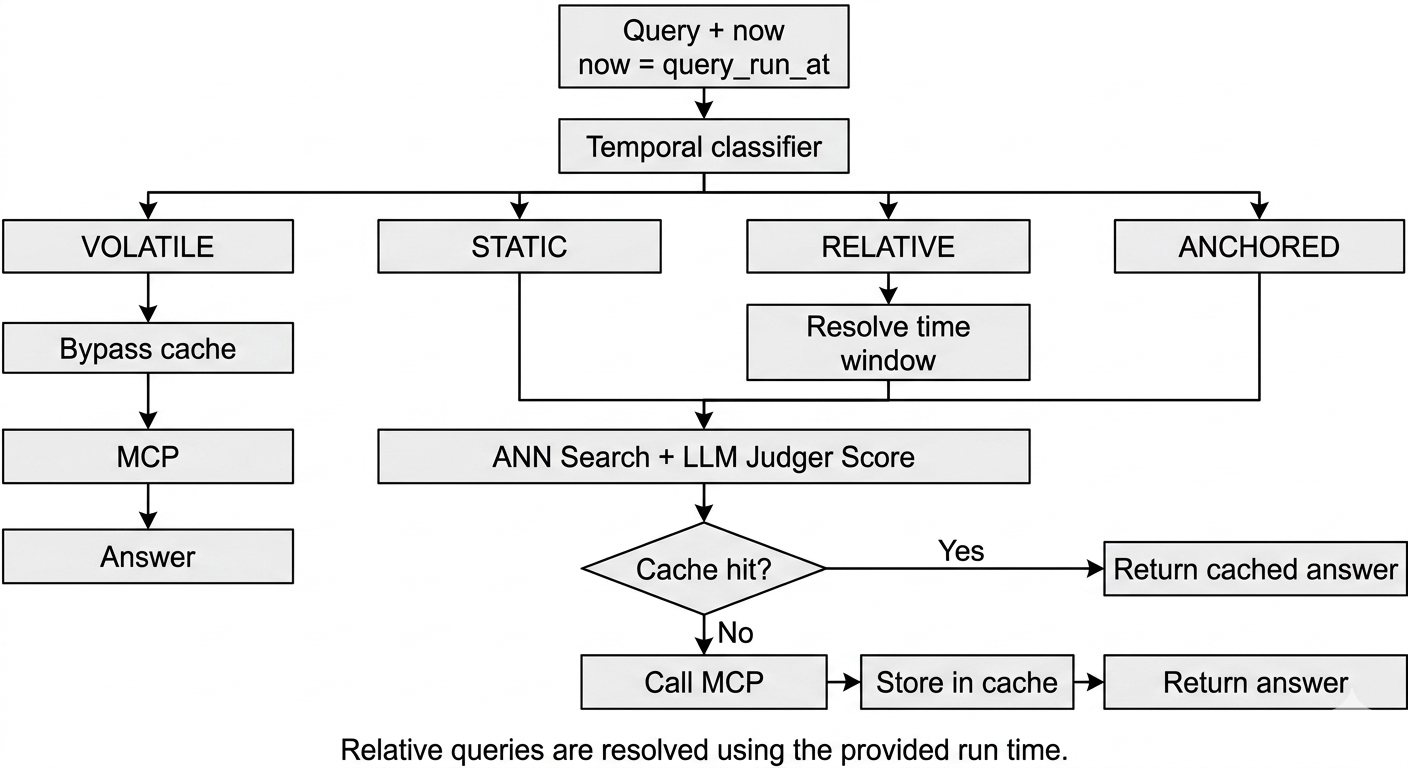}
    \caption{Temporal semantic cache workflow. A pre-retrieval temporal classifier routes each query: Volatile bypasses the cache; Static and resolved-Anchored queries enter ANN retrieval followed by a reranker-based judger.}
    \label{fig:temporal-semantic-cache}
\end{figure}

Figure~\ref{fig:temporal-semantic-cache} shows the temporal semantic cache. Each incoming query is paired with a run-time timestamp and passed through a temporal classifier before semantic retrieval. The classifier assigns each query to one of four buckets. \textit{Volatile} queries request live system state and bypass the cache. \textit{Static} queries have no temporal dependence and enter semantic retrieval. \textit{Relative} queries use expressions like ``yesterday'' or ``last week'' that are resolved against the run timestamp into concrete windows and then treated as Anchored. \textit{Anchored queries} reference a fixed time window and enter approximate nearest-neighbor retrieval with a window-aware judger. On a hit, the cached answer is returned; on a miss, the query falls through to the full plan-execute pipeline, and the resulting answer is inserted into the cache. The MCP layer (Figure~\ref{fig:mcp-workflow}) sits inside this pipeline: even on a miss, discovery caching and parallel execution reduce wall-clock cost relative to the unoptimized baseline.

\subsection{Cache Layer Design Choices}

\paragraph{Why temporal classification?}
A naive semantic cache would embed every query and search the index. This conflates linguistic similarity with answer reuse validity, as discussed in Section~\ref{sec:caching-limits}. Placing a temporal filter \emph{before} retrieval lets us route each query into a regime where reuse is sound: live-state queries skip the cache entirely, time-bounded queries match only against compatible windows, and time-independent queries fall back to standard semantic retrieval. The classifier itself is a lightweight regex-based component that adds negligible per-query cost.

\paragraph{Anchored windows over relative phrases.}
Relative time expressions like ``yesterday'' resolve differently each day, so caching them under their literal text would produce stale hits. We resolve such phrases against the query timestamp at insertion time and store the concrete window with the cache entry. At lookup, the judger checks window compatibility as part of its acceptance decision.

\paragraph{Embedding plus reranker, not similarity alone.}
Cosine similarity over query embeddings is a coarse signal. We therefore use it only for candidate retrieval and route candidates through a reranker-based judger that scores semantic and temporal alignment with the new query. This two-stage design~\citep{schroeder2025vectorq} lets us tune retrieval recall and judging precision independently. Exact thresholds and model choices are listed in Appendix~\ref{app:impl}.

\subsection{MCP Workflow Optimizations}

\paragraph{Discovery-phase caching.}
The baseline AOB pipeline performs MCP tool discovery on every query: spawning a Python subprocess for each of the four servers, establishing \texttt{stdio} connections, requesting the tool catalog via \texttt{list\_tools()}, and terminating before planning begins. In our setup this consumes 2 to 3 seconds per query. We treat tool signatures as semi-static metadata and persist the aggregated catalog to a local JSON file. The cache key invalidates automatically on changes to server source code, server registrations, or project configuration; full key construction is in Appendix~\ref{app:impl}. The left side of Figure \ref{fig:mcp-workflow-grid} illustrates how we designed this caching mechanism.

\paragraph{Parallel step execution.}
We treat the generated plan as a directed acyclic graph
of tool invocations and group steps into topological
dependency layers. Independent steps within a layer
execute concurrently, and dependency barriers preserve
ordering across layers. To support concurrent execution,
a persistent \texttt{MCPServerPool} maintains one
\texttt{stdio} session per required server for the
lifetime of a plan, with per-server asynchronous locks
serializing concurrent calls to the same domain server
while allowing inter-server concurrency. The executor
is fail-tolerant: a failure on one MCP server does not
halt sibling steps targeting other servers. The right side of Figure \ref{fig:mcp-workflow-grid} illustrates how we set up the parralel step execution over the persistent server pool.

\begin{figure}[ht]
    \centering
    
    \begin{subfigure}{0.48\linewidth}
        \centering
        \includegraphics[width=\linewidth]{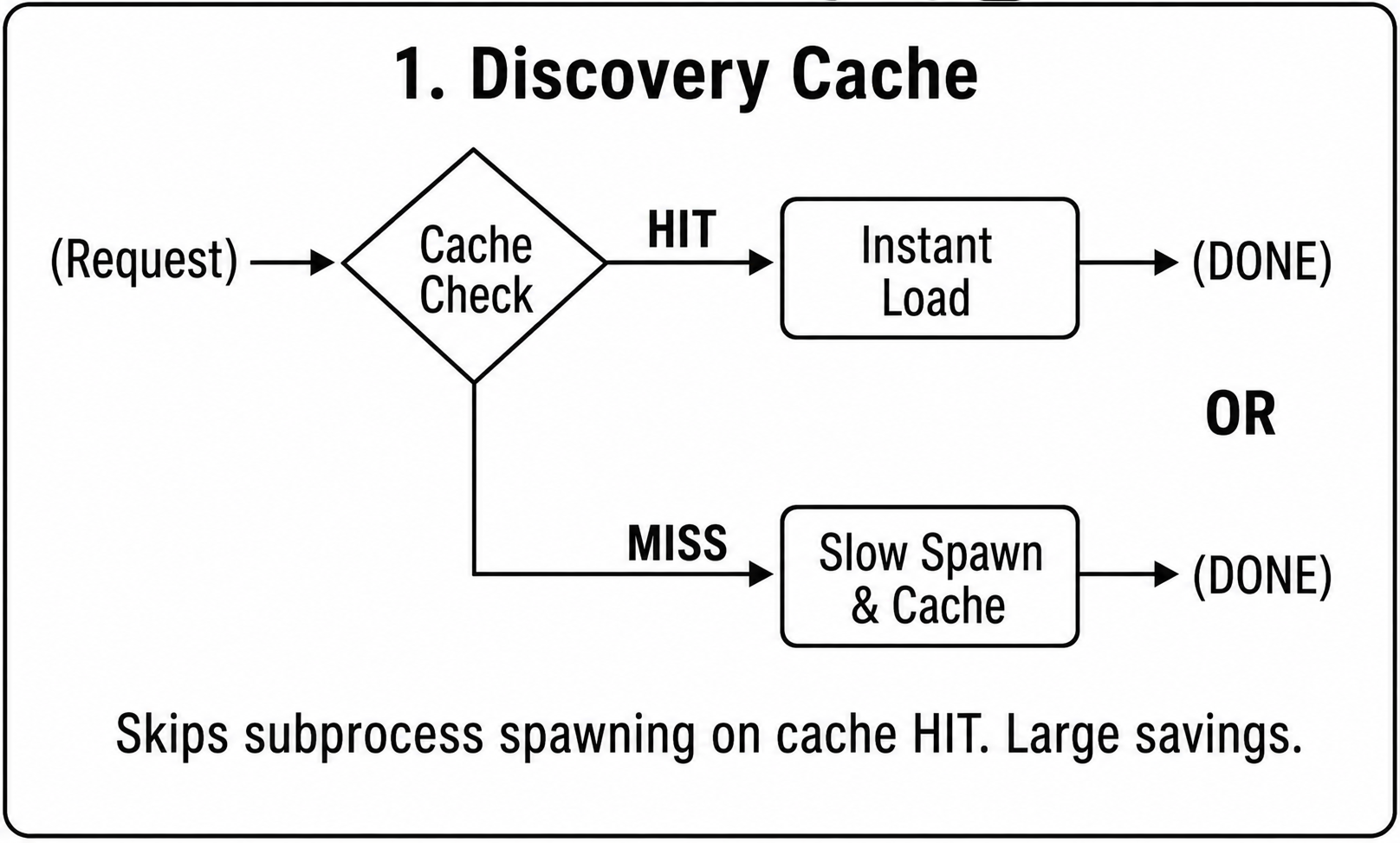}
    \end{subfigure}
    \hfill
    \begin{subfigure}{0.48\linewidth}
        \centering
        \includegraphics[width=\linewidth]{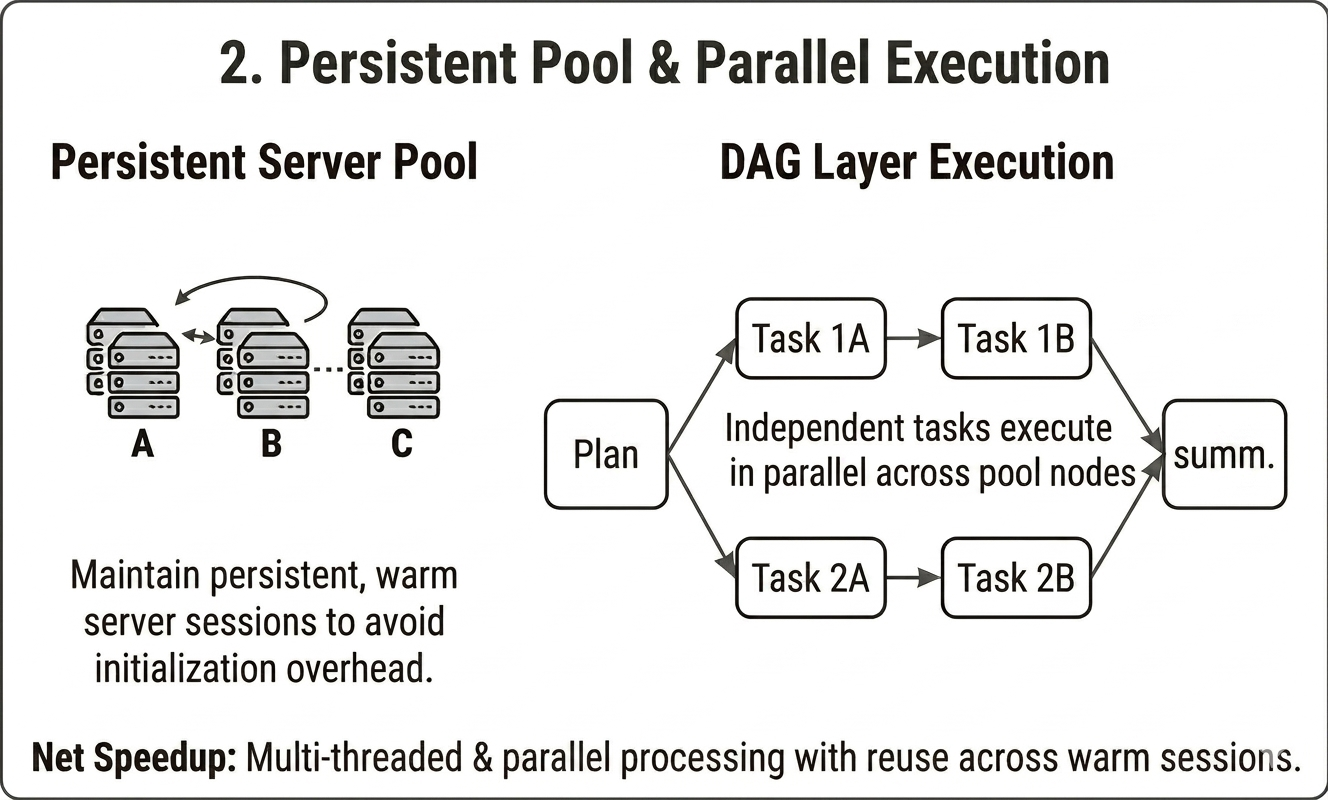}
    \end{subfigure}

    \caption{ The optimized MCP Workflow component paths use a discovery cache and dispatch steps in parallel against a persistent pool.}
    \label{fig:mcp-workflow-grid}
\end{figure}

\section{Results and Evaluation}

We evaluate the framework on AOB queries and report the following key findings:
\begin{itemize}
    \item \textbf{End-to-end speedup.} The combined pipeline reduces median latency from 34.10s to 9.80s ($3.48\times$) on 80 paraphrase-tier queries (Section~\ref{sec:results-combined}).
    \item \textbf{MCP workflow gains.} On 18 IoT queries, MCP optimizations alone yield a $1.67\times$ end-to-end speedup, with discovery cost reduced by $296\times$ and execution time by $1.99\times$ (Section~\ref{sec:results-mcp}).
    \item \textbf{Cache decision quality.} The temporal-classifier-plus-judger reaches F1 0.64 on hit/miss decisions in the combined system (Section~\ref{sec:results-combined}), with the residual error concentrated on parameter-shifted queries.
    \item \textbf{Additive optimizations.} The miss path remains faster than the unoptimized baseline in our experiments because MCP gains apply regardless of cache state.
\end{itemize}

\subsection{Experiment Setup}
\label{sec:results-setup}

\paragraph{Benchmark workload.}
All scenarios are drawn from \texttt{all\_utterance.csv}, the hand-authored AOB corpus of 152 queries spanning IoT, FMSR, TSFM, Work Order, and multi-agent types. Because the two optimization layers target different latency sources, we use two purposive subsets. For the \emph{MCP-workflow scenarios}, we ran the AOB planner over the corpus and retained 20 queries whose plans contained at least two parallelizable branches; this subset is intended as a parallelism stress test (not a representative slice of the full corpus). For the \emph{cache scenarios}, we randomly partition parents into 20 \emph{warm} parents and a held-out cold pool. We then use an LLM to generate semantically similar query paraphrases for each parent query, emit one paraphrase per warm parent as a 20-row seed CSV, and emit an 80-row test CSV split 60\%/40\% between warm-parent paraphrases (cache should hit) and cold-parent paraphrases (cache should miss). The \texttt{parent\_id} membership in the warm set serves as ground truth for hit/miss labelling.

\paragraph{Baselines.}
For the workflow experiment, the baseline performs tool discovery on every query and executes plan steps sequentially. For the cache experiment, the baseline is the workflow-optimized pipeline with cache lookup and insert disabled, isolating the cache's contribution from the workflow contribution. Each query is run paired under baseline and optimized conditions on the same simulated wall-clock so that row-level latency differences are attributable to the optimization rather than provider-side variance.

\paragraph{Metrics.}
We report per-query end-to-end latency with the median as primary statistic and the 5\%-trimmed mean as a robustness check. Speedups are reported as the median of per-row ratios, robust to provider-side tail-latency events. For the workflow experiment, we additionally break out per-phase latency. For each test row the cache emits a \emph{hit} or \emph{miss}; pairing this with the \texttt{parent\_id} ground truth produces a $2\times2$ confusion matrix from which we compute precision, recall, F$_1$, and specificity. For misses we report median overhead (cached minus baseline latency).

\paragraph{Implementation.}
The plan-execute pipeline uses Llama-3.3-70B via LiteLLM for planning, tool-argument resolution, and summarization~\citep{llama3, litellm}. The semantic cache uses Qwen3 embedding and reranker models with FAISS-based ANN retrieval~\citep{qwen3, faiss}. All experiments run on a single Apple M-series machine with 16~GB unified memory. Exact model strings, threshold values, cache capacity, and hardware are listed in Appendix~\ref{app:impl}.

\subsection{MCP Workflow Optimization Results}
\label{sec:results-mcp}

We evaluate the MCP layer in isolation on the 18 IoT queries from the AOB benchmark, each executed three times in baseline and optimized configurations (120 total profiled runs). Two queries (Q5, Q19) timed out across all attempts in both modes and are excluded.

\begin{table}[h]
\centering
\caption{Aggregate phase-level comparison across 18 IoT queries (median of per-query medians, 3 runs each).}
\label{tab:aggregate}
\begin{tabular}{lrrrr}
\toprule
\textbf{Phase} & \textbf{Baseline} & \textbf{Optimized} & \textbf{Saving} & \textbf{Speedup} \\
\midrule
Discovery       & 2.096s  & 0.007s  & 2.089s  & $296.08\times$ \\
Planning (LLM)  & 10.285s & 8.226s  & 2.059s  & $1.25\times$ \\
Pre-fetch       & 0.415s  & 0.002s  & 0.413s  & $210.53\times$ \\
Execution       & 34.639s & 17.415s & 17.224s & $1.99\times$ \\
Summarization   & 5.016s  & 5.165s  & $-$0.149s & $0.97\times$ \\
\midrule
\textbf{Total}  & \textbf{56.902s} & \textbf{34.164s} & \textbf{22.738s} & \textbf{$1.67\times$} \\
\bottomrule
\end{tabular}
\end{table}

\paragraph{Phase-level effects are surgical.}
Discovery caching effectively eliminates per-query server-spawning overhead. Parallel execution with connection pooling reduces execution wall time by $1.99\times$. Planning and summarization, both dominated by LLM inference, show no statistically significant change, confirming that the optimizations target only the orchestration layer and do not introduce overhead in untargeted phases. The combined effect is a $1.67\times$ end-to-end median speedup with an average saving of 22.7 seconds per query (40\% reduction).

\paragraph{Per-query gains correlate with parallelism.}
The optimized pipeline achieves greater than $1.0\times$ on 16 of 18 queries, with the largest gains on plans that have multiple independent branches (Q16: $5.06\times$, Q3: $3.27\times$, Q6: $3.03\times$). Two queries show modest regression (Q1: $0.92\times$, Q11: $0.67\times$); both regressions trace to LLM-side variance in summarization rather than overhead from the optimizations. Appendix~\ref{app:perquery} provides the full per-query distribution and a worked structural comparison for Q6.

\subsection{End-to-End Combined Pipeline}
\label{sec:results-combined}

We now evaluate the full pipeline (cache + MCP optimizations) versus the unoptimized baseline on 80 paraphrase-tier queries derived from the 20 AOB IoT seed scenarios.

\paragraph{Latency.}
The baseline achieves a median end-to-end latency of 34.10s (mean 68.68s, range 6.73s to 398.73s). The fully optimized pipeline reduces this to 9.80s (mean 33.06s, range 0.26s to 230.78s), an overall $3.48\times$ median speedup. The cache hit rate is 45.0\% (36 of 80 rows). On hit rows the optimized pipeline bypasses plan-execute entirely and returns a cached response, yielding $31.87\times$ median speedup and saving a median of 25.50s per row.

\begin{figure}[h]
    \centering
    \includegraphics[width=0.62\linewidth, keepaspectratio]{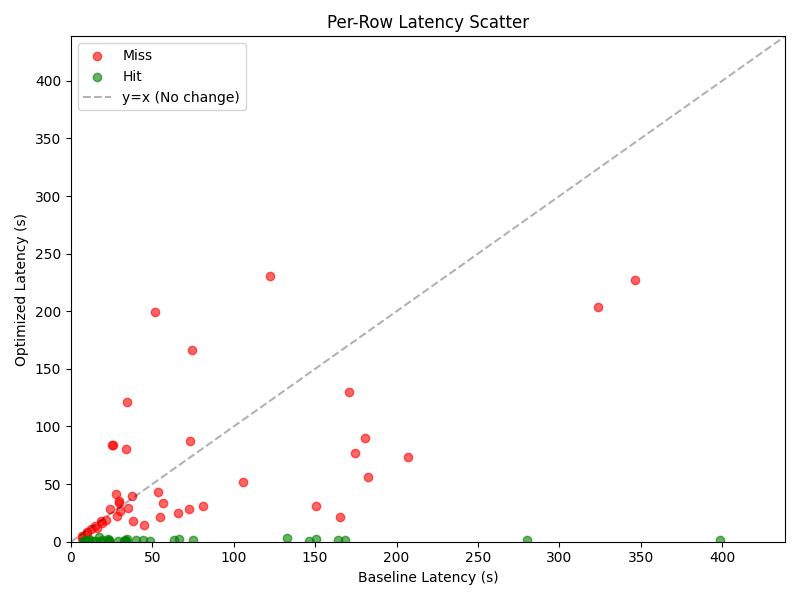}
    \caption{Per-row latency for all 80 evaluation queries. Cache hits collapse to near-zero optimized latency; misses track below the baseline because MCP optimizations apply regardless of cache state.}
    \label{fig:per-row-latency}
\end{figure}

\paragraph{Miss path is still faster.}
On the 44 miss rows the optimized pipeline still beats the baseline: the median latency difference is $-3.30$s. This saving comes from the MCP layer alone. The miss path therefore incurs no net overhead relative to the unoptimized baseline; the cache lookup cost is more than recovered by MCP-side gains. Figure~\ref{fig:per-row-latency} visualizes this: hit rows collapse to near-zero optimized latency, miss rows track below the baseline.

\paragraph{Cache decision quality and the parameter-sensitivity ceiling.}
On the combined pipeline the cache reaches precision 0.75, recall 0.5625, F1 0.6429, and specificity 0.7188. Compared with the cache-only configuration in Appendix~\ref{app:cache-asteria}, precision improves (0.667 to 0.75) while recall drops slightly, reflecting a more conservative judger when the full pipeline is available as fallback. The residual errors concentrate on parameter-shifted queries: paraphrases that differ only in asset ID or sensor name embed close to seed entries, pass the similarity gate, and require the judger to make a fine-grained distinction that the embedding does not surface. This empirically caps F1 in our setting and motivates parameter-aware judging as future work.

\paragraph{The two layers are additive.}
The MCP optimizations provide a consistent latency reduction on every query regardless of cache state, while temporal semantic caching adds a large further reduction on the subset of queries that resolve to valid hits. Neither layer undermines the other: the MCP-optimized miss path is faster than the unoptimized baseline, so the cache never makes miss-path performance worse.

\section{Limitations and Failure Modes}
\label{sec:limitations}

The most important limit of our approach is structural rather than incidental: pure semantic similarity is not a sound proxy for answer validity in parameter-rich industrial queries, and no choice of similarity threshold can fully resolve this. We discuss this and other limits below.

\paragraph{Parameter-collision false positives.}
The dominant failure mode is cross-parameter false positives. Queries that share linguistic frame but differ in asset, sensor, or time window can embed at cosine similarity above 0.95, then pass the reranker-based judge above the strict $\tau_{\text{judge}}=0.92$ threshold, and return an answer drawn from a different operational context. Concretely: in our evaluation we observed a query asking for ``Tonnage sensor readings for Chiller~6 \emph{and} Chiller~9 at MAIN site during the first week of \emph{December 2020}'' return a cached answer originally produced for ``\% Loaded data for Chiller~6 at MAIN site for \emph{June 2020},'' with a judger score of 0.97. The two queries differ in three operational dimensions (sensor, equipment scope, time window), but the embedding is dominated by their shared frame (``sensor readings for Chiller~X at MAIN site''). This pattern caps hit-decision F1 near 0.64 in the combined system (Section~\ref{sec:results-combined}) and accounts for the 0.5625 recall: tightening $\tau_{\text{judge}}$ further would suppress these false positives but at the cost of additional missed legitimate paraphrases. Within pure semantic caching, this is unfixable; Section~\ref{sec:future} discusses parameter-aware extensions.

\paragraph{Judger inconsistency on legitimate paraphrases.}
A separate failure mode is judger noise on clear paraphrases. We observed paraphrases of the same seed query receiving judger scores from $0.5$ to $0.95$ on semantically equivalent content, with no obvious pattern in the residual error. We attribute this to the bounded capacity of \texttt{Qwen3-Reranker-0.6B}, which is the smallest variant of its family. Larger or domain-adapted rerankers may reduce this variance.

\paragraph{Workload structure caps the achievable hit rate.}
The 80-row paraphrase tier in Section~\ref{sec:results-combined} is biased toward warm-parent paraphrases by construction (60\%/40\% warm/cold split), giving a hit rate of 45\%. On the full unfiltered AOB corpus the achievable hit rate is structurally lower, between roughly 15\% and 30\% in our setup, because most AOB queries are parameter-rich data fetches (specific asset IDs, specific sensors, specific time windows) where pure semantic matching is fundamentally unsafe. The setting in which our cache contributes most cleanly is knowledge-style queries (failure-mode enumeration, sensor-to-component mappings, model-support questions), which are a subset of the AOB workload.

\paragraph{Excluded queries and provider variance.}
Two of the 20 IoT queries used for the MCP workflow experiment (Q5 and Q19) timed out across all attempts in both baseline and optimized configurations and are excluded from Table~\ref{tab:aggregate} and Figure~\ref{fig:perquery}. These timeouts are upstream LLM failures rather than orchestration failures, but they are worth flagging as a scope limit on the per-query analysis. Additionally, the two MCP regressions in Figure~\ref{fig:perquery} (Q1 at $0.92\times$, Q11 at $0.67\times$) trace to summarization-phase variance under WatsonX provider load: Q11's summarization took 18.6s in one optimized run versus 7.9s in baseline. This is a property of the LLM endpoint rather than our optimizations, but it does mean per-query speedups have a noise floor we cannot eliminate.

\paragraph{Window grammar and natural-date handling.}
The temporal classifier resolves a fixed grammar of relative phrases (``yesterday,'' ``last week,'' explicit ISO ranges) into concrete windows. Natural-date phrases like ``June 2020'' or ``the last week of December 2020'' currently extract as bucket=Anchored but window=None, so they are demoted to the Static path at lookup. This works correctly but loses the temporal-prefilter benefit for a non-trivial fraction of AOB queries. A richer date parser would promote these into proper Anchored buckets.

\paragraph{Single-machine evaluation, no persistence.}
All experiments run on a single Apple M-series machine with 16~GB unified memory. Inter-machine variability and concurrent-load effects are out of scope. The cache also lives only in memory: a process restart loses all warmed entries, and there is no checkpoint or replay mechanism. Both are addressed in future work.

\section{Future Work}
\label{sec:future}

The limitations above point to several concrete extensions, ordered roughly from most committed to most exploratory.

\paragraph{Parameter-aware caching.}
The natural follow-up to the parameter-collision finding is to extract structured parameters (entity, sensor, time window, action verb) from each query and cache under a key of the form (canonical\_intent, param\_combo). A parameter-exact lookup short-circuits to the cached answer; semantic matching only fires when parameter sets overlap. This combines hash-keyed precision with paraphrase-robustness and would eliminate the false-positive class observed in Section~\ref{sec:limitations}.

\paragraph{Hybrid retrieval at the lookup layer.}
A lightweight parameter-extraction layer sits in front of the existing temporal-and-semantic cache. The classifier outputs both a temporal bucket (as today) and a parameter signature; lookup first attempts an exact-parameter match, then falls back to semantic retrieval restricted to entries with overlapping parameters. This integrates naturally with the existing Asteria substrate.

\paragraph{A larger or domain-adapted reranker.}
Replacing \texttt{Qwen3-Reranker-0.6B} with the 4B variant or with a reranker fine-tuned on AOB-style query-answer pairs may reduce the judger inconsistency described in Section~\ref{sec:limitations}.

\paragraph{Richer natural-date grammar.}
A date parser that handles ``June 2020,'' ``Sept 19 2020 at 7pm,'' and similar natural expressions would promote currently-demoted Anchored queries back into the temporal pre-filter path, improving cache hit safety on time-bounded queries.

\paragraph{Cache persistence.}
Pickling the cache state plus a FAISS index round-trip with versioning would avoid the 30-second pre-warm cost on every process restart. This is mechanical engineering work but is necessary for any non-trivial deployment.

\paragraph{Online judge threshold recalibration.}
Asteria~\citep{asteria2025} specifies an online ground-truth-sampling procedure for $\tau_{\text{judge}}$ tuning. Our implementation currently uses a single offline-set threshold; an online recalibration loop would adapt to workload drift.

\paragraph{Scaling the evaluation.}
The 152-utterance AOB corpus is small. Generating 1000+ utterances using the existing paraphrase generator, with stratification across query types and parameter shifts, would tighten confidence in the failure-mode and speedup claims.

\paragraph{Integration with serving infrastructure.}
The temporal cache and MCP optimizations are orthogonal to engine-level optimizations such as PagedAttention~\citep{kwon2023vllm} or SGLang's structured-program execution~\citep{zheng2024sglang}. Combining the two layers in a production deployment is a separate engineering exercise but should compound the gains reported here.

\section{Conclusion}
We presented two complementary optimization layers for AOB plan-execute pipelines: a temporal semantic cache that classifies queries by time-sensitivity before semantic retrieval, and a set of MCP workflow optimizations that eliminate per-query discovery overhead and parallelize independent plan steps. The two layers are additive: MCP optimizations reduce latency on every query regardless of cache state, while temporal semantic caching adds large further savings on the subset of queries that resolve to valid hits. On 18 IoT queries the MCP layer alone yields a $1.67\times$ end-to-end speedup; on the 80-row paraphrase tier the combined pipeline yields $3.48\times$. In our experiments, the miss path remains faster than the unoptimized baseline.

Beyond the speedup, the contribution we want reviewers to weigh is the failure-mode analysis. Pure semantic similarity, even paired with a strict reranker-based judge, is not a sound proxy for answer validity in parameter-rich industrial queries: shared linguistic frame dominates the embedding, distinct operational parameters do not. Our hit-decision F1 caps near $0.64$ in this setting, with the residual error concentrated on cross-parameter false positives that pass even at $\tau_{\text{judge}}=0.92$. This is a structural property of pure semantic caching as an evaluation/optimization choice for MCP-backed industrial agents, not a tuning issue, and it gives a concrete handle on when caching is safe to deploy as part of an evaluation pipeline. Parameter-aware caching, the natural next step, is laid out in Section~\ref{sec:future}.

\section*{Acknowledgments}
We thank Dhaval Patel and the IBM team for providing access to AssetOpsBench and for their guidance throughout this work. We thank Kaoutar El Maghraoui for her instruction and mentorship over the course of this project.

\bibliographystyle{unsrtnat}
\bibliography{references}

\newpage
\appendix

\section{Implementation Parameters}
\label{app:impl}

\paragraph{Discovery cache.}
The cache key is computed as an MD5 hash over three components: the registered server paths, the last-modified timestamps (\texttt{mtime}) of all Python source files within the \texttt{src/servers/} directories, and the modification time of the repository's \texttt{pyproject.toml} dependency file. Any change to server logic or project dependencies invalidates the key automatically. We additionally enforce a 24-hour time-to-live (TTL).

\paragraph{Parallel executor.}
The executor uses Kahn's algorithm to group plan steps into topological dependency layers and dispatches all steps within a layer concurrently using \texttt{asyncio.gather()}. The \texttt{MCPServerPool} maintains one persistent \texttt{stdio} session per required server, with per-server asynchronous locks serializing concurrent calls to the same server.

\paragraph{Semantic cache models and thresholds.}
Embeddings are produced by \texttt{Qwen3-Embedding-0.6B} (1024-dim). The judger is \texttt{Qwen3-Reranker-0.6B} run with prefill-only inference. Both models run on Apple Silicon MPS in fp16. ANN retrieval uses FAISS with \texttt{top\_k=5} and a coarse cosine threshold of \texttt{tau\_sim=0.75}. The judger applies a strict acceptance threshold of \texttt{tau\_jsm=0.92}. Cache capacity is 50 entries with LCFU eviction.

\paragraph{LLM backend.}
Planning, tool-argument resolution, and summarization use \texttt{watsonx/meta-llama/llama-3-3-70b-instruct} via LiteLLM.

\paragraph{Hardware.}
All experiments run on a single Apple M-series machine with 16~GB unified memory.

\section{Cache-Only Configuration: Decision Breakdown}
\label{app:cache-asteria}

\begin{figure}[H]
    \centering
    \begin{minipage}{0.42\linewidth}
        \centering
        \includegraphics[width=\linewidth, keepaspectratio]{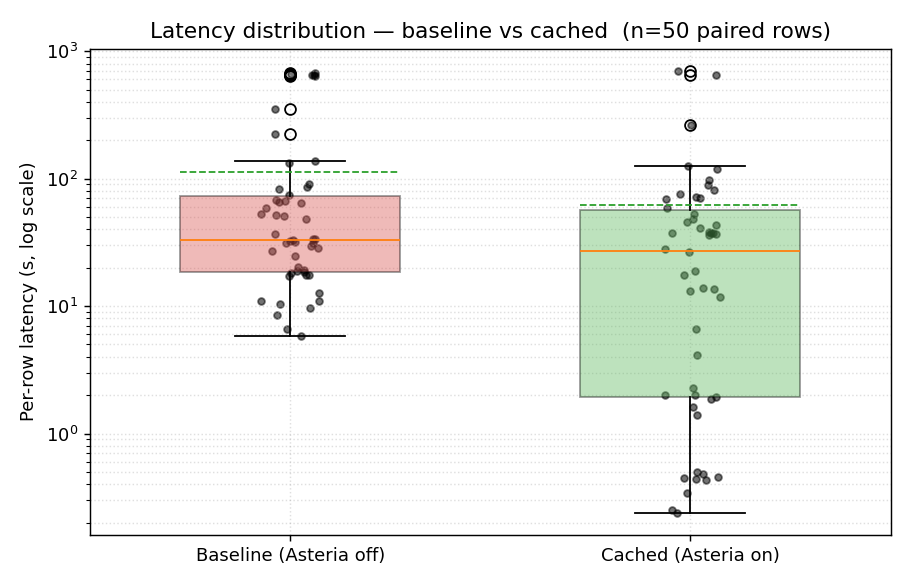}
        \caption{Box plot of baseline and cached latency distributions across the 50 evaluation rows.}
        \label{fig:asteria-boxplot}
        \end{minipage}\hfill
    \begin{minipage}{0.55\linewidth}
        \centering
        \includegraphics[width=\linewidth, keepaspectratio]{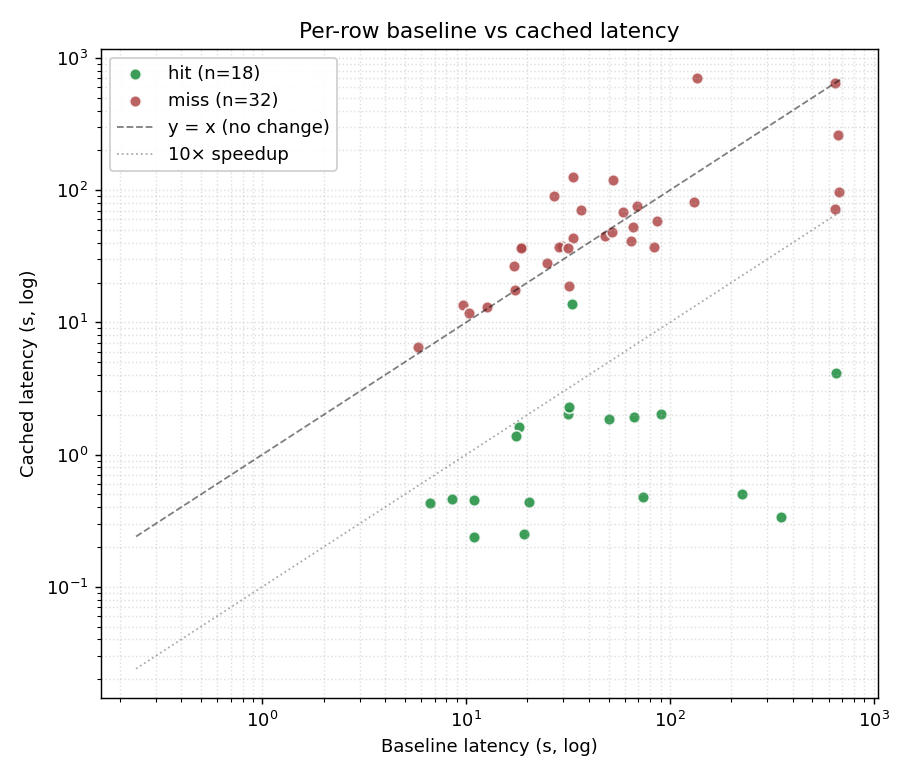}
        \caption{Cache-only per-row latency. Hits collapse to near-zero cached cost regardless of baseline.}
        \label{fig:asteria-scatter}
    \end{minipage}
\end{figure}

We additionally evaluate the temporal semantic cache as a standalone layer on top of the unmodified plan-execute pipeline (no MCP optimizations) on a stratified sample of 50 queries. The cache achieves a hit rate of 36.0\%, with median speedup $30.62\times$ on hits and median overhead $+2.23$s on misses. Decision quality reaches precision 0.667, recall 0.667, F1 0.667, specificity 0.813. Figure~\ref{fig:asteria-boxplot} shows boxplot for asteria optimization; Figure~\ref{fig:asteria-scatter} shows per-row latency.

\section{Per-Query Speedup and Structural Comparison}
\label{app:perquery}

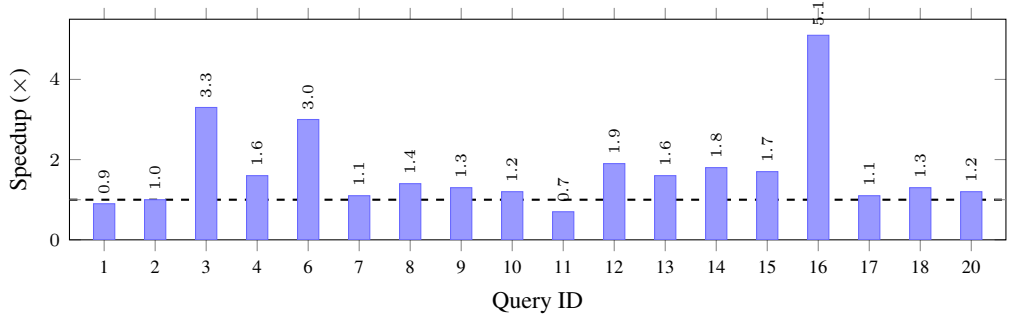
\begin{figure}[H]
\centering
\begin{tikzpicture}
\begin{axis}[
    ybar,
    width=\textwidth,
    height=4.5cm,
    bar width=8pt,
    ylabel={Speedup ($\times$)},
    xlabel={Query ID},
    symbolic x coords={1,2,3,4,6,7,8,9,10,11,12,13,14,15,16,17,18,20},
    xtick=data,
    x tick label style={font=\scriptsize},
    y tick label style={font=\scriptsize},
    ylabel style={font=\small},
    xlabel style={font=\small},
    ymin=0, ymax=5.5,
    extra y ticks={1},
    extra y tick style={grid=major, grid style={black, thick, dashed}},
    extra y tick labels={},
    nodes near coords,
    nodes near coords style={font=\tiny, rotate=90, anchor=west},
    every node near coord/.append style={/pgf/number format/.cd, fixed, fixed zerofill, precision=1},
    enlarge x limits=0.04,
]
\addplot[fill=blue!40, draw=blue!60] coordinates {
    (1,0.9) (2,1.0) (3,3.3) (4,1.6) (6,3.0) (7,1.1)
    (8,1.4) (9,1.3) (10,1.2) (11,0.7) (12,1.9) (13,1.6)
    (14,1.8) (15,1.7) (16,5.1) (17,1.1) (18,1.3) (20,1.2)
};
\end{axis}
\end{tikzpicture}
\caption{Per-query end-to-end speedup across 18 completed IoT queries. Dashed line marks $1.0\times$ break-even.}
\label{fig:perquery}
\end{figure}

\begin{figure}[H]
\centering
\begin{tikzpicture}[
    node distance=0.35cm and 0.4cm,
    sbox/.style={rectangle, draw, rounded corners=2pt, minimum height=0.55cm, minimum width=1.6cm, align=center, font=\scriptsize, fill=red!8},
    obox/.style={rectangle, draw, rounded corners=2pt, minimum height=0.55cm, minimum width=1.6cm, align=center, font=\scriptsize, fill=green!10},
    lbl/.style={font=\scriptsize\bfseries, align=center},
    arr/.style={-{Stealth[length=1.5mm]}, thick},
    timenode/.style={font=\scriptsize\itshape, text=gray},
]
\node[lbl] (bl) {Baseline (Sequential)};
\node[sbox, below=0.3cm of bl] (bd) {Discovery\\2.45s};
\node[sbox, right=0.3cm of bd] (bp) {Plan\\9.66s};
\node[sbox, right=0.3cm of bp] (bs1) {Step 1\\};
\node[sbox, right=0.3cm of bs1] (bs2) {Step 2};
\node[sbox, right=0.3cm of bs2] (bs3) {Step 3};
\node[sbox, right=0.3cm of bs3] (bs4) {Step 4};
\node[sbox, right=0.3cm of bs4] (bs5) {Step 5};
\node[sbox, right=0.3cm of bs5] (bsum) {Summ.\\4.17s};

\draw[arr] (bd) -- (bp);
\draw[arr] (bp) -- (bs1);
\draw[arr] (bs1) -- (bs2);
\draw[arr] (bs2) -- (bs3);
\draw[arr] (bs3) -- (bs4);
\draw[arr] (bs4) -- (bs5);
\draw[arr] (bs5) -- (bsum);

\node[timenode, right=0.2cm of bsum] {152.9s};

\node[obox, below=0.9cm of bd] (od) {Cache hit\\0.008s};
\node[lbl, above=0.2cm of od.north west, anchor=south west] (ol) {Optimized (Parallel + Cache + Pool)};

\node[obox, right=0.3cm of od] (op) {Plan\\12.37s};

\node[obox, right=0.6cm of op, yshift=0.3cm] (os1) {Step 1};
\node[obox, right=0.6cm of op, yshift=-0.3cm] (os2) {Step 2};

\node[obox, right=0.3cm of os1, yshift=0cm] (os3) {Step 3};
\node[obox, right=0.3cm of os2, yshift=0cm] (os4) {Step 4};

\node[obox, right=0.6cm of os3, yshift=-0.3cm] (os5) {Step 5};

\node[obox, right=0.3cm of os5] (osum) {Summ.\\5.07s};

\draw[arr] (od) -- (op);
\draw[arr] (op) -- (os1);
\draw[arr] (op) -- (os2);
\draw[arr] (os1) -- (os3);
\draw[arr] (os2) -- (os4);
\draw[arr] (os3) -- (os5);
\draw[arr] (os4) -- (os5);
\draw[arr] (os5) -- (osum);

\node[timenode, right=0.2cm of osum] {50.5s};

\end{tikzpicture}
\caption{Workflow comparison for Q6. Top: baseline sequential execution with subprocess-per-call. Bottom: optimized execution with discovery cache, parallel DAG layers, and persistent server pool.}
\label{fig:workflow}
\end{figure}
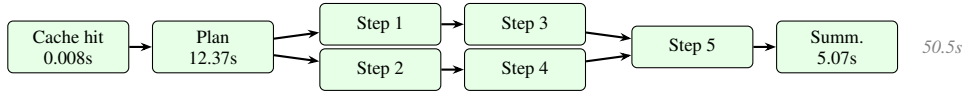

Figure~\ref{fig:workflow} shows a worked structural comparison for Query~6 (failure modes of Chiller~6 detectable by its Chiller Efficiency sensor; 5-step plan, two dependency layers). In the baseline path each step spawns a fresh MCP subprocess, executes a single tool call, and terminates before the next step begins. In the optimized path the server pool starts each required server once, the discovery cache bypasses spawning entirely, and independent steps within each dependency layer execute concurrently. Total latency drops from 152.9s to 50.5s, a $3.03\times$ speedup.

\section{Broader Impact and Societal Implications}
\label{app:broader-impact}

\paragraph{Positive impacts.}
Reducing per-query latency and API cost makes LLM-backed agent systems more accessible to organizations that cannot afford high-throughput commercial serving. In industrial operations, faster and cheaper query resolution can improve response times for maintenance workflows, equipment fault detection, and work-order management, with downstream benefits for operational safety and efficiency. The workflow optimizations are backend-agnostic and can compound with engine-level improvements such as PagedAttention, so their benefits extend beyond the specific AOB setting studied here.

\paragraph{Potential negative impacts.}
The primary risk introduced by any caching layer is stale or incorrect answer reuse. In safety-critical industrial settings (e.g., returning a cached sensor reading that no longer reflects current equipment state), an incorrect cache hit could inform a faulty maintenance decision. Our temporal classifier mitigates this by routing live-state queries past the cache, but it does not eliminate the risk entirely, particularly for Anchored queries whose window grammar does not parse correctly. Practitioners deploying this system in safety-critical contexts should treat cache hits as advisory and maintain a fallback to live query execution. No direct path to misuse for disinformation, surveillance, or discriminatory decision-making is introduced by this work.

\end{document}